\documentclass[conference,letterpaper]{ieeeconf}
\IEEEoverridecommandlockouts

\usepackage{ifpdf}

\usepackage{cite}

  \usepackage[pdftex]{graphicx}
  \graphicspath{{./figures/}}
  \DeclareGraphicsExtensions{.pdf,.jpeg,.png}
\usepackage{tikz}
\usetikzlibrary{external}
\usetikzlibrary{arrows}
\usepackage{pgfplots}
\pgfplotsset{compat=1.5}
\pgfplotsset{every tick label/.append style={font=\fontsize{7}{8}\selectfont},
every axis legend/.append style={font=\fontsize{7}{8}\selectfont},
every axis label/.append style={font=\fontsize{8}{9}\selectfont}}

\usepackage{amsfonts,amssymb}
\usepackage[cmex10]{amsmath}
\usepackage{bm}

\usepackage[ruled,linesnumbered,vlined]{algorithm2e}
\usepackage{algpseudocode} 

\usepackage{array}

\usepackage{mdwmath}
\usepackage{mdwtab}

\usepackage{multirow}

\usepackage[caption=false]{subfig}

\usepackage{stfloats}

\usepackage{url}

\newcommand*{\policy}{\pi}
\newcommand*{\action}{a}
\newcommand*{\ActionSpace}{\mathcal{A}}
\newcommand*{\reward}{R}

\newcommand*{\argmax}[1]{\underset{#1}{\mathrm{argmax}~}}
\newcommand*{\argmin}[1]{\underset{#1}{\mathrm{argmin}~}}

\newcommand*{\weight}{w}
\newcommand*{\weights}{{\mathbf{w}}}
\newcommand*{\wSet}{W}
\newcommand*{\tp}{x}
\newcommand*{\TpSpace}{\mathcal{X}}
\newcommand*{\tpSet}{X}
\newcommand*{\ip}{\hat{x}}
\newcommand{\ips}{\hat{X}}
\newcommand*{\transpose}{\mathrm{T}}
\newcommand*{\R}{\mathbb{R}}
\newcommand*{\Hspace}{\mathcal{H}}
\newcommand*{\feature}{\bm{\hat{\phi}}}
\newcommand*{\featureDim}{M}

\newcommand*{\methodName}{Coordinate Descent Bayesian Optimisation}
\newcommand*{\methodShortName}{CDBO}
\newcommand*{\algorithmName}{StochasticCoordinateAscent}

\begin{document}
%
\title{Learning to Race through \methodName{}}

\author{Rafael Oliveira, Fernando H.M. Rocha, Lionel Ott, Vitor Guizilini, Fabio Ramos and Valdir Grassi Jr.
\thanks{Rafael Oliveira, Lionel Ott, Vitor Guizilini and Fabio Ramos are with the School of Information Technologies, The University of Sydney, Australia.
\texttt{rdos6788@uni.sydney.edu.au}
\newline
Fernando H.M. Rocha and Valdir Grassi Jr. are with the Sao Carlos School of Engineering at the University of Sao Paulo, Sao Carlos, SP, Brazil,
\texttt{fernandorocha@usp.br}
\newline
This work was supported by \emph{Coordena\c{c}\~{a}o de
Aperfei\c{c}oamento de Pessoal de N\'{i}vel Superior} (CAPES), Brazil, and Data61/CSIRO, Australia.
}
}



\maketitle

\begin{abstract}
In the automation of many kinds of processes, the observable outcome can often be described as the combined effect of an entire sequence of actions, or controls, applied throughout its execution. In these cases, strategies to optimise control policies for individual stages of the process might not be applicable, and instead the whole policy might have to be optimised at once. On the other hand, the cost to evaluate the policy's performance might also be high, being desirable that a solution can be found with as few interactions as possible with the real system. We consider the problem of optimising control policies to allow a robot to complete a given race track within a minimum amount of time. We assume that the robot has no prior information about the track or its own dynamical model, just an initial valid driving example. Localisation is only applied to monitor the robot and to provide an indication of its position along the track's centre axis. We propose a method for finding a policy that minimises the time per lap while keeping the vehicle on the track using a Bayesian optimisation (BO) approach over a reproducing kernel Hilbert space. We apply an algorithm to search more efficiently over high-dimensional policy-parameter spaces with BO, by iterating over each dimension individually, in a sequential coordinate descent-like scheme. Experiments demonstrate the performance of the algorithm against other methods in a simulated car racing environment.
\end{abstract}


\section{Introduction}
In the automation of various kinds of physical systems, sometimes a controller, or a control policy, needs to be optimised based only on a total cost or reward evaluating its performance. In robotics, the physical processes of interest are often related to the motion or navigation of a robot. In some cases, it is impossible to quantify the effect that control actions at individual steps have on the final outcome of the process. Examples include tasks such as ball-throwing \cite{Kupcsik2013} or trying to hit a target on a wall with a dart \cite{Kober-RSS-10}. In other cases, individual rewards might be quantifiable for each step taken, but a global approach, considering a whole episode of execution, could yield better results, such as in autonomous racing \cite{Rizano2013}.

In the case of autonomous racing, the problem of finding a control policy that will allow a robot racer to finish a track in minimal time has been approached in many ways. Model-predictive control (MPC) techniques have been applied to locally optimise driving policies over receding-horizons based on external sensor data and internal dynamical models, which are either pre-designed \cite{Liniger2015} or learnt \cite{Williams2017}. 
From a global optimisation perspective, the same problem has been approached as racing line optimisation \cite{Rizano2013}, i.e. finding the path within the track that will allow the car to finish the race in minimal time, given a map of the race track and a kinematic model of the vehicle. When global information about the racing environment is available through external images, deep reinforcement learning architectures have also been applied to fine-tune control policies \cite{Sallab2017}. Lastly, this problem has also been approached by using evolutionary algorithms to optimise parameters for policies that either combine different pre-designed heuristics \cite{Butz2009}, or directly map the state of the car and the opponents to control actions via some type of network structure \cite{Sanchez2014}.

In this paper, we are concerned with the problem of enabling a robot to learn the control policy that will allow it to complete a lap in a given race track as fast as possible, improving over a single initial example given by a sub-optimal controller. This is performed without the need for a model of the robot or a map of the track, but by performing multiple laps and learning from the outcome of each of them. Model-based approaches are generally limited by the ability of the model to represent the behaviour of the robot \cite{Deisenroth2013}, which can be hard to capture for real robots. In addition, these approaches usually require the Markov assumption to be valid for the model representation, which implies that enough information about the dynamic state of the robot is observable.

In a policy-search framework, ideally we look for policies which are flexible enough to represent a variety of behaviours and an optimisation procedure that can find the best policy within a minimal amount of episodes and requires only minimal information about the system. Radial basis function (RBF) networks \cite{Deisenroth2013} are among the classes of policy parameterisations with high representation power. These policies, however, can be very high dimensional and challenging to optimise with conventional optimisation algorithms within a limited budget of policy evaluations. Bayesian optimisation (BO) \cite{Brochu2010}, on the other hand, is a strategy to optimise expensive-to-evaluate functions within a limited budget of function evaluations that has shown promising results in policy search problems \cite{Martinez-Cantin2007,Wilson2014}. Although methods which apply BO in high-dimensional search spaces have been proposed \cite{Wang2013,LiChun2016}, it still remains an open question how to do so without making restrictive assumptions about the objective function.

Our contribution is a method that applies Bayesian optimisation (BO) \cite{Brochu2010} to optimise control policies with high-dimensional parameter spaces. In each iteration of the BO loop, our method sequentially optimises each parameter of the policy in turn, following a randomised coordinate descent \cite{Nesterov2012} scheme. This is performed over a Gaussian process \cite{Rasmussen2006} surrogate, modelling a reward function, that is sequentially updated. The method is applied to optimise control policies for a racing car in a race simulator. We employ a general class of control policies, defined by feature expansions in a reproducing kernel Hilbert space (RKHS) \cite{Scholkopf2002}. RBF networks can be seen as a specialised instance of this generic representation. 

The remainder of this paper is organised as follows. In the next section, we review relevant related work in the area of Bayesian optimisation. In Section \ref{sec:prel}, we formally present the optimisation problem approached by this paper and some background information about BO. Then, in Section \ref{sec:method},  we present our method. In Section \ref{sec:exp}, we present experimental evaluations of the method using a race car simulator, comparing the proposed method against other optimisation algorithms. Finally, in Section \ref{sec:conclusion}, we conclude and propose some directions for future work.

\section{Related Work}
\label{sec:related}
In this section, we review relevant prior work in the areas of reinforcement learning and Bayesian optimisation.

Learning a control policy that maximises a reward that depends on the combined effect of its actions can also be approached as episode-based policy search in reinforcement learning (RL) \cite{Deisenroth2013}. In \cite{Kupcsik2013}, for example, a framework is proposed to allow algorithms that learn policies using an internal, also learned, forward model of the robot to simulate trajectory roll-outs. In \cite{Kober-RSS-10}, a robot learns how to hit a target throwing a dart using a model-free approach.
This algorithm models the policy parameters distribution with a Gaussian process (GP) \cite{Rasmussen2006} prior over the contexts space for each single parameter. 

Bayesian optimisation \cite{Brochu2010} has also been applied to problems involving policy search. A similar approach to the one in this paper is presented in \cite{Martinez-Cantin2007}, where the authors model cost as a direct function of policy parameters. In that paper, the robot's task was to reduce uncertainty about its location and its surroundings, and the policies were parameterised by a small number of variables. Since the policy-execution outcome/reward is usually more-closely a function of the resulting behaviour of the robot than of the policy parameters, another approach to this problem is applying a GP prior over this mapping from behaviours to rewards \cite{Wilson2014}. This approach, however, usually requires that enough information can be observed from the system. Our aim in this paper is to learn policies that can have enough representation power, which generally implies high-dimensional parameter spaces, while requiring the least amount of information from the system.

One common issue with the presented applications of BO is the curse of dimensionality. The great majority of BO algorithms uses GPs to learn and model the objective function, which does not scale well with high dimensions and/or large amounts of data, degrading the performance of BO in such settings. Several approaches have been recently developed to tackle the scaling of GP/BO to high dimensions. One could, for example, assume that the objective function only depends on a small subset of input coordinates and do variable selection to find these \cite{Chen2012,Ulmasov2016}. Another approach is to assume there is a lower-dimensional linear embedding containing most of the variation of the function, including its optimum, and using linear projections to reduce the dimensionality of the search space \cite{Wang2013,Djolonga2013}. Another common approach is to assume that the function is formed by a set of low dimensional disjoint functions \cite{LiChun2016} combined in an additive structure, which allows learning separate GP models and optimising over them separately.

Given the particular nature of our problem, we chose to tackle issues that arise in high-dimensional BO by approaching it from the search side, while still using all available information to build a model of the objective function. In problems like racing, due to the constraints imposed by the environment, i.e. the track, and the dynamic limits on the agent, i.e. the car, we end up having a reward function whose mass is highly concentrated within a particular region of the search space. Therefore, we argue that, by applying a relatively simple method such as randomised coordinate descent (CD) \cite{Wright2015}, and starting the search from a valid initial solution, we can optimise over a high-dimensional GP model efficiently. CD relies on the fact that each sub-problem is a lower-dimensional optimisation problem, which is solved more easily than the full problem. We empirically demonstrate that this simple search strategy when combined with BO can be effective in the particular class of problems we approach.

\section{Preliminaries}
\label{sec:prel}
In this section, we present some preliminary information for the work in this paper. We start with a formal description of the particular problem we are dealing with, followed by the formulation of our policies parametrisation. After that, we review some basic concepts in Bayesian optimisation and Gaussian process regression.

\subsection{Problem Statement}
\label{sec:problem}
Consider policies mapping the position $\tp \in \TpSpace$ of the robot along the track to a corresponding control action $\action \in \ActionSpace$. Both $\TpSpace$ and $\ActionSpace$ are continuous spaces. Our goal is to find the control policy $\policy: \TpSpace \to \ActionSpace$ that minimises the time $T$ required to complete the track. 

We utilise a reward $\reward$ defined as:
\begin{equation}
\reward =
\begin{cases}
L/T, ~\text{if track completed}\\
0, ~\text{if failed to complete the track}
\end{cases}
\end{equation}
where $L$ is the length of the track. Therefore, for success cases, the reward is equivalent to the average linear speed of the robot. In this sense, minimising $T$ is equivalent to maximising $\reward$. With that, we search for:
\begin{equation}
\policy^* = \argmax{\policy \in \ActionSpace^\TpSpace} ~ \reward[\policy]
\label{eq:problem}.
\end{equation}

We seek a method that solves the above problem for any kind of mobile robot and without needing a map of the track. Therefore, model-based solutions are out of scope, for they would have to learn an approximate transition model of the robot, whose representation varies among different driving mechanisms, and use this model to simulate trajectories over the track map.

\subsection{Policy Parameterisation}
\label{sec:policy}
We assume that the optimal $\policy$ belongs to a reproducing kernel Hilbert space (RKHS) \cite{Scholkopf2002}. This approach has been relatively successful in modelling trajectories for motion planning \cite{Marinho2016} and components of stochastic control policies for reinforcement learning \cite{Bagnell2003,Vien2016}. This formulation allows the policy to assume a variety of shapes, depending on the choice of kernel function, allowing one to encode prior knowledge about the ideal control policy. Considering a 1-D action space $\ActionSpace \subset \mathbb{R}$, a policy can be represented by:
\begin{equation}
\policy(\tp) = \sum_{i=1}^N  \alpha_i k(\tp_i,\tp),
\label{eq:policy}
\end{equation}
where $k: \TpSpace \times \TpSpace \to \R$ is a positive-definite kernel function, and $\alpha_i \in \R$ and $\tp_i \in \TpSpace$ are arbitrary. Besides that, it is possible that $N \to \infty$.

In a RKHS, kernels are equivalent to inner products between features mappings $\phi: \TpSpace \to \mathcal{H}$ in the corresponding Hilbert space, such that $k(\tp,\tp') = \langle \phi(\tp), \phi(\tp')\rangle_\Hspace$. With that, $\policy(\tp)$ can also be represented by $\policy(\tp) = \sum_{i=1}^{N} \alpha_i \langle \phi(\tp_i), \phi(\tp) \rangle_\Hspace$. By the linearity property of inner products, we can move the sum inside, and have:
\begin{equation}
\policy(\tp) = \langle \sum_{i=1}^{N} \alpha_i \phi(\tp_i), \phi(\tp) \rangle_\Hspace = \langle \weight, \phi(\tp) \rangle,
\end{equation}
where $\weight \in \Hspace$, which can be infinite in dimensions. Several techniques in the kernel machines literature, e.g. \cite{Le2013}, however, propose approximating $\phi(\tp)$ by a vector $\bm{\hat{\phi}}(\tp) \in \R^\featureDim$, $\featureDim < \infty$, such that $\bm{\hat{\phi}}(\tp)^\transpose \bm{\hat{\phi}}(\tp') \approx k(\tp,\tp')$. As a consequence,
\begin{equation}
\policy(\tp) \approx \policy_\weights (\tp) = \weights^\transpose \bm{\hat{\phi}}(\tp),
\end{equation}
where $\weights \in \R^\featureDim$ is a vector of scalar weights, which uniquely determines the policy $\policy$ for a given feature mapping $\feature$. Therefore, we can rewrite Equation \ref{eq:problem} as:
\begin{equation}
\weights^* = \argmax{\weights \in \R^\featureDim} R[\policy_\weights].
\end{equation}

In our case, the advantage of the features approximation is that we don't need to deal with the individual $\alpha_i$'s and $\tp_i$'s, which can be many more than $\featureDim$, but only with their combined effect on the control policy. More importantly, we also reduce the dimensionality of the search space, projecting it to $\featureDim$ dimensions. 

Although the problem has been formulated for 1-D actions, it could be easily extended to actions composed by multiple independent controls, by simultaneously optimising the weight vectors for each corresponding policy. The drawback, however, is the multiplication of the number of weights to optimise.


\subsection{Bayesian Optimisation}
We use BO \cite{Brochu2010} to perform the policy weights optimisation as it allows finding the global optimum of arbitrary functions that are expensive to evaluate. To aid in that, BO applies a Bayesian model, which is typically a Gaussian process (GP) \cite{Rasmussen2006}, as a prior to internally approximate the objective function. 

At each iteration, BO selects the point to perform the next evaluation of the objective by maximising an acquisition function over the model. This acquisition function provides a utility value that enables the algorithm to perform a guided search for the global optimum, and is usually much simpler to evaluate than the objective function. It represents a natural trade-off between exploration (searching for areas with high uncertainty) and exploitation (searching for areas where the objective function is expected to be high), and aims to minimise the number of objective function evaluations.  

After evaluating the objective function at the selected point, the prior is updated with the new observation and the algorithm proceeds to the next iteration, keeping track of the current optimum. As iterations proceed, it can be shown that the prior approximation converges to the objective function within the given search space, and consequently the global optimum of the objective can be found.

\subsection{Gaussian Processes}
Gaussian process \cite{Rasmussen2006} regression is a Bayesian non-parametric framework that places a Gaussian distribution as a prior over the space of functions mapping the inputs $\weights \in \mathbb{R}^\featureDim$ to outputs $z \in \mathbb{R}$, where $z = f(\weights) + \epsilon$ is a noisy observation of the true underlying reward value $f(\weights)$, and $\epsilon \sim \mathcal{N}(0,\sigma_{n}^2)$ is Gaussian-distributed noise with zero mean and standard deviation $\sigma_{n}$. A GP model can be completely specified by a mean and a covariance function, $k_\reward$, which is a positive-definite kernel. Using zero as the mean for the prior, the values of $f$ for a set of $Q$ points $\wSet^* = \{\weights^*_i \in \mathbb{R}^\featureDim\}_{i=1}^Q$ obey a multivariate normal distribution:
\begin{equation}
\mathbf{f}^* = f(\wSet^*) \sim \mathcal{N}(\mathbf{0},k_\reward(\wSet^*,\wSet^*)),
\end{equation}
where $f(\wSet^*) = [f(\weights^*_1),\dots,f(\weights^*_Q)]^\transpose$, and $k_\reward(\wSet^*,\wSet^*)$ is an $Q$-by-$Q$ matrix whose elements are determined by $k_\reward(\wSet^*,\wSet^*)_{ij} = k_\reward(\weights^*_i,\weights^*_j)$. Given a set $\{\wSet,\mathbf{z}\}$ of $N$ observations of $f$, where $\wSet = \{\weights_i \in \mathbb{R}^\featureDim\}_{i=1}^{N}$ and $\mathbf{z} = \{z_i \in \mathbb{R}\}_{i=1}^N$, the joint distribution of the observed outputs and the function values at the query points under the GP prior is given by:
\begin{equation}
\left[
\begin{array}{c}
\mathbf{z}\\
\mathbf{f}^*
\end{array}
\right]
\sim
\mathcal{N}
\left(
\mathbf{0},
\left[
\begin{array}{cc}
k_\reward(\wSet,\wSet) + \sigma_n^2I & k_\reward(\wSet,\wSet^*)\\
k_\reward(\wSet^*,\wSet) & k_\reward(\wSet^*,\wSet^*)
\end{array}
\right]
\right)
\end{equation}
Conditioning this joint on the observations, inference in a GP can be done by:
\begin{equation}
\mathbf{f}^*|\wSet,\mathbf{z},\wSet^* \sim \mathcal{N}(\bm{\mu}^*,\bm{\Sigma}^*),
\end{equation}
where:
\begin{align}
\bm{\mu}^* &= k_\reward(\wSet^*,\wSet)K_\wSet^{-1}\mathbf{z}\\
\bm{\Sigma}^* &= k_\reward(\wSet^*,\wSet^*) - {k_\reward}(\wSet^*,\wSet)K_\wSet^{-1}{k_\reward}(\wSet,\wSet^*),
\end{align}
using ${K_\wSet} = k_\reward(\wSet,\wSet) + \sigma_n^2I$, with each $k_\reward(\wSet,\wSet)_{ij} = k_\reward(\weights_i,\weights_j)$.

\section{\methodName{}}
\label{sec:method}
We approach the problem in Section \ref{sec:problem} from a Bayesian optimisation (BO) perspective, which places a Gaussian process (GP) prior over the objective function, in our case, $\reward[\policy_\weights] = f(\weights)$, and optimises it by doing searches over the GP-based surrogate model. 

\subsection{Acquisition Function Optimisation} 
In this paper, another problem that we face is the possibly high dimensionality of the search space for the optimisation of the acquisition function (AF) that BO utilises. Quite a few methods have been proposed in the BO literature to deal with high-dimensional search spaces, such as \cite{Wang2013,LiChun2016}. However, they usually require a few possibly strong assumptions about the objective function. We instead employ a very simple method, Stochastic Coordinate Ascent (Algorithm \ref{alg:axissearch}), which uses a random axis selection scheme to optimise the AF over each axis individually, starting from the current optimum candidate. This way, we bias the search locally, avoiding excessive exploration in a high-dimensional space. Depending on the choice of acquisition function, however, we can still perform a global search with BO by exploring regions of high uncertainty around the current optimum location.

Our AF optimisation technique was inspired by Coordinate Descent (CD), a class of algorithms that are one of the oldest in the optimisation literature \cite{Rosenbrock1960}. It is based on the idea that an $n$-dimensional problem can be decomposed in $n$ one-dimensional sub-problems, which makes it suitable to be applied to large or high-dimensional datasets \cite{Nesterov2012}. In the CD strategy, each coordinate is updated sequentially by solving the one-dimension problem with any suitable optimisation algorithm, while all the others dimensions are kept fixed. The methods vary in the way the sequence of dimensions is chosen, if it optimises only one dimension or a block, and if it uses gradients on the iterations or not. Several works focused on describing the convergence characteristics of these algorithms \cite{Nesterov1983, Bertsekas2015,Nesterov2012}. Under some assumptions (e.g. Lipschitz continuity, strong convexity) they prove a linear convergence rate for the sequential and randomised CD method, and also for the block-CD.
The most intuitive scheme for this technique would be to optimise the dimensions in an ordered cyclical fashion. But \cite{Powell1973} showed an example of non-convex function that, when applied this sequential scheme, the optimisation process cycles without converging. To avoid these kind of issues, in our work we adopt a randomised sequence that changes once every dimension has been optimised.

\begin{algorithm}
\caption{\algorithmName{}}
\label{alg:axissearch}
\KwIn{$h, \weights^*, \mathcal{D}$}
$\weights = \weights^*$\;
$\mathcal{I} = \mathtt{randomShuffle}(\{1,\dots,\featureDim\})$\;
\For{$i \in \mathcal{I}$}
{
	$\weight_i = \argmax{v \in \R}{h(\weight_0,\dots,\weight_{i-1},v,\weight_{i+1},\dots,\weight_\featureDim|\mathcal{D})}$\;
}
\Return $\weights$
\end{algorithm}

Algorithm \ref{alg:axissearch} starts the search from the current optimum location. It randomly shuffles a sequence of $\featureDim$ dimension indices $\{1,\dots,\featureDim\}$, and follows the shuffled sequence to optimise one axis after the other. After passing through all the dimensions, it returns the corresponding optimised vector of weights.

Although not using derivatives, as in standard coordinate descent methods \cite{Nesterov2012}, in practice our approach is still able to achieve similar results. We performed preliminary tests applying other CD methods, e.g. the randomised accelerated coordinate descent method (RACDM) \cite{Nesterov2012}. We observed that their performance is usually not as good as the simple scheme in Algorithm \ref{alg:axissearch}. Reasons for that involve issues with escaping saddle points and restrictions in the GP covariance function, which needs to be differentiable for the method to be applicable. The latter is not the case of Mat\'ern 1 \cite{Rasmussen2006}, for example, the best performing in our experiments.

\subsection{The Policy-Search Algorithm} 
\label{sec:algo}
We propose an algorithm that improves a valid initial policy so that the robot racer can finish the track in minimal time. The initial policy can be obtained by recording the actions of a simple controller or a human driver, obtaining an initial set of points $\tpSet = \{\tp_i\}_{i=1}^n$ and the corresponding observed actions $\mathbf{\action}=\{\action_i\}_{i=1}^n$. The initial weights $\weights_0$ can then be fitted by minimising the quadratic cost function:
\begin{equation}
\weights_0 = \argmin{\weights \in \R^\featureDim} \lVert\mathbf{\action}-\feature(\tpSet)\weights\rVert_2^2 + \lambda \lVert\weights\rVert_2^2,
\end{equation}
where $\feature(\tpSet) = [\feature(\tp_1),\dots,\feature(\tp_n)]^\transpose$ is a matrix with the features for each $\tp_i$ on the corresponding row and $\lambda$ is a regularisation factor, to avoid extreme values for the weights. A solution to this problem can be found analytically by zeroing out the gradient of the fit term with respect to $\weights$, yielding:
\begin{equation}
\weights_0 = [\feature(\tpSet)^\transpose\feature(\tpSet)+\lambda I]^{-1}\feature(\tpSet)^\transpose \mathbf{\action}.
\end{equation}

Before starting the policy search with BO, an initial training of the GP is needed to provide an estimate of its hyper-parameters, which are the noise variance and the parameters of the covariance function in our case. Due to the vast majority of the search space being, in general, composed of invalid policies, using uniform or Latin hyper-cube random samples could provide too many observations with zero reward. This can cause over-fitting to the initial GP hyper-parameter selection. To avoid this, we sample and execute a set of $S$ samples from a normal distribution $\mathcal{N}(\weights_0,I\sigma_0^2)$ to form an initial dataset $\{\weights_i, \reward_i\}_{i=1}^{S}$ to feed the GP with, so that BO can have an informative prior over the reward function. In addition, the hyper-parameters can also be re-estimated online after each observation of a reward.

The proposed method is summarised in Algorithm \ref{alg:main}. In lines 1 through 4, it collects the initial set of observations for the GP. In lines 5 through 12, the search for the optimal policy is performed. Line 7 executes the policy parameterised by the current weights. Lines 8 through 10 keep track of the optimum. Line 11 updates the GP dataset. Line 12 performs the maximisation of the acquisition function to select the next vector of weights to evaluate. The algorithm proceeds until its budget of $N$ function evaluations is exhausted.

\begin{algorithm}
\caption{\methodShortName}
\label{alg:main}
\DontPrintSemicolon
\KwIn{
$\weights_0$: weights of the initial policy\linebreak
$\sigma_0^2$: initial samples variance\linebreak
$S$: number of initial samples\linebreak
$N$: number of laps}

\For {$s=1\dots S$}
{
	Sample $\weights_s \sim \mathcal{N}(\weights_0,I\sigma_0^2)$\;
	$\reward_s \leftarrow$ Execute $\policy(\tp;\weights_s)$\;
}

$\mathcal{D} \leftarrow \{\weights_s,\reward_s\}_{s=1}^S$\;
$\reward^* \leftarrow \max \reward_{s=1,\dots,S}$\;

\For {$i=0\dots N-S$}
{
	$\reward_i \leftarrow$ Execute $\policy_{\weights_i}$\; \label{line:loop_start}
	\If{\label{line:optimum1}$\reward_i > \reward^*$}
	{
		$\reward^* \leftarrow \reward_i$\;
		$\weights^* \leftarrow \weights_i$\;
	}
	$\mathcal{D} \leftarrow \mathcal{D} \cup \{(\weights_i,\reward_i)\}$\label{line:gp1_update}\;
	$\weights_{i+1} = \mathtt{\algorithmName{}}(h,\weights^*,\mathcal{D})$\; 
}
\Return $\weights^*, \reward^*$
\end{algorithm}



\section{Experiments}
\label{sec:exp}
In this section, we present the experimental evaluation of the performance of our method in simulation. We performed tests with a robot car driving on race tracks performing realistic full physics simulation using the race engine of an open-source game, called \textit{Speed Dreams}\footnote{Speed Dreams: https://sourceforge.net/projects/speed-dreams/}, which is based on TORCS \cite{TORCS}. In all the tests, we compared the performance of the algorithm against:
\begin{itemize}
\item CMA-ES, which has been applied to reinforcement learning problems \cite{Rucks2010}, in particular, we use active CMA-ES\cite{Arnold2010}, with an implementation provided by an open-source library\footnote{\url{https://github.com/beniz/libcmaes}};
\item standard BO, using CMA-ES to optimise the acquisition function.
\item REMBO \cite{Wang2013}, a Bayesian Optimisation technique that uses random embeddings, developed to deal with high-dimensional problems. We tested with both 5 and 10-dimensional random embeddings.
\end{itemize}
To run the internal optimisation along each coordinate in our method, we used COBYLA \cite{Powell2007}, a local derivative-free optimisation algorithm, with an implementation provided by a popular non-linear optimisation library \cite{NLopt}.

\subsection{Setup}
The state space of the robot is represented as its position along a given race line normalised by its length, i.e. $\tp \in [0,1]$, where 0 corresponds to the start line, increasing to 1 when the robot crosses the finish line. As race line, we used the centre line of the track, but it could be any other valid trajectory, allowing our method to be combined with racing line optimisation algorithms. The control policy actuates the car's acceleration by optimising throttle and braking, which are combined into a single scalar output $\action \in [-1,1]$, with positive values for throttle and negative for braking. The steering control was performed using a simple proportional-integral (PI) controller, which tries to minimise the distance between the car and the racing line. We have not approached the steering control in this paper, since optimising only the acceleration control of the car is already a challenging problem and sufficient to demonstrate the capabilities of the proposed BO method in dealing with high dimensions.

\begin{figure}
\centering
\includegraphics[width=0.7\columnwidth]{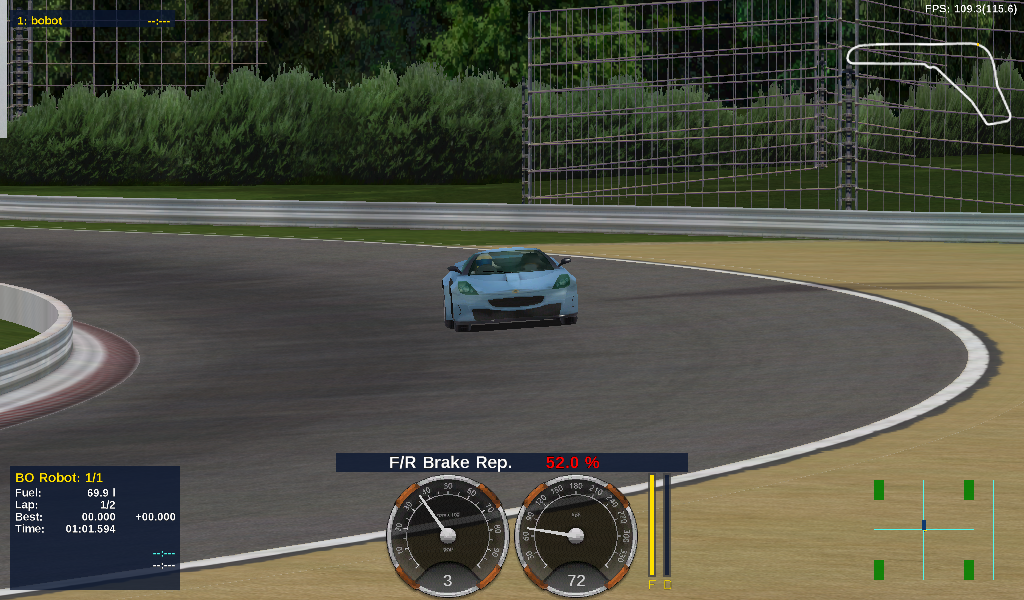}
\caption{Screen-shot of the race car used by our algorithm in simulations using the game engine.}
\label{fig:car}
\end{figure}

In this setup, the algorithm needed to be cautious not to set huge acceleration values for the car at critical parts of the track, like curves, both to not destabilise the steering controller and also to respect the friction limits of the tires. Figure \ref{fig:car} presents a screen-shot with the car model we used, a \textit{Spirit 300}.

For features $\feature$, we utilised an array of $\featureDim$ kernels placed over a set of inducing points along the track, i.e.:
\begin{equation}
\feature(\tp) = [k_\action(\tp,\ip_1),\dots,k_\action(\tp,\ip_\featureDim)]^\transpose.
\end{equation}
We chose $\ips = \{\frac{i}{\featureDim-1}\}_{i=0}^{\featureDim-1}$, so that it forms a set of $\featureDim$ regularly-spaced points along the track. In this sense, no prior information about where the critical parts of the track are was assumed. However, a possible way to choose $\ips$ would be to place points around portions of the track requiring a significant change in acceleration, such as curves, which would require a map of the track. This could also be learnt by analysing the initial control policy, searching for areas of high variation.

\begin{figure} 
	\tikzsetnextfilename{lengthscales}
	\begin{tikzpicture}
	\begin{axis}[
	xlabel={$\tp$},
	ylabel={$\action$},
	y=6cm,ymax=0.3,ymin=-0.05,xmin=0.2,xmax=0.6,x=18cm,
	legend style={at={(0.98,0.95)},anchor=north east}, legend columns=4
	]
	\addplot[black,dashed,thick] table[header=false,x index=0,y index=1, each nth point=50, filter discard warning=false, unbounded coords=discard, skip first n=200] {data/forza/lengthscales-effect/initial-actions.dat};
	\addlegendentry{recorded}
	\addplot[red,solid,thick] table[header=false,x index=0,y index=1, skip first n=200, each nth point=2, filter discard warning=false, unbounded coords=discard] {data/forza/lengthscales-effect/policy-ls0.01.dat};
	\addlegendentry{$l=0.01$}
	\addplot[blue,solid,thick] table[header=false,x index=0,y index=1, skip first n=200, each nth point=5, filter discard warning=false, unbounded coords=discard] {data/forza/lengthscales-effect/policy-ls0.05.dat};
	\addlegendentry{$l=0.05$}
	\addplot[green,solid,thick] table[header=false,x index=0,y index=1, skip first n=100, each nth point=10, filter discard warning=false, unbounded coords=discard] {data/forza/lengthscales-effect/policy-ls0.2.dat};
	\addlegendentry{$l=0.2$}
	\end{axis}
	\end{tikzpicture}
\caption{Detail of the effect of different lengths scales on the fitting of the initial policy for the same kernel placement. The recorded actions are shown in a dashed line. Shorter length scales allow sharper transition, but might compromise interpolation. Longer length scales yield smoother curves, but might compromise flexibility}
\label{fig:lengthscales}
\end{figure}
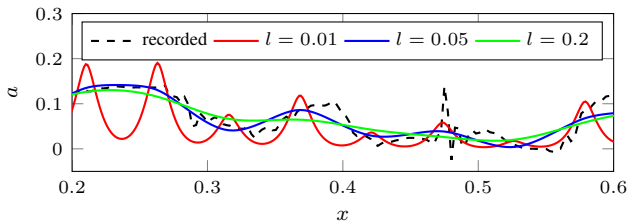

If we set $k_\action$ to be the RBF squared exponential kernel, we have the RBF network policy parameterisation, which has already been applied to some reinforcement learning problems \cite{Deisenroth2013}. However, that kernel has very smooth transitions, and would not be flexible enough to provide fast transitions in the commands profile. Therefore, for all the experiments, we used the \textit{Mat{\'e}rn} class of kernel functions (see \cite{Rasmussen2006}, Chapter 4) for the policies, $k_\action$, in particular, we used Mat{\'e}rn 3:
\begin{equation}
k_\action(\tp,\tp') = \left(1 + \frac{\sqrt{3}}{l}|\tp-\tp'|\right)\exp\left(-\frac{\sqrt{3}}{l}|\tp-\tp'|\right),
\end{equation}
where $l$ is a length-scale parameter controlling the smoothness of the curve. We set $l$ with values around the spacing of the inducing points, i.e. $l \approx \frac{1}{\featureDim-1}$, in our experiments so that the gaps between the kernels can be filled without compromising the flexibility of the function. Figure \ref{fig:lengthscales} demonstrates the effect that different length scales have on the shape of the curve. The fitting of the length scale can also be numerically optimised before starting the race. 

For the GP model, we used the Mat{\'e}rn 1 kernel as covariance function, which allows us some flexibility to model sharp transitions in the reward function. This Mat{\'e}rn kernel is equivalent to the exponential covariance function:
\begin{equation}
k_\reward(\weights,\weights') = \sigma_f^2\exp\left(-\sqrt{d^2(\weights,\weights')}\right),
\end{equation}
where $\sigma_f^2$ is a signal variance parameter and $d^2(\weight,\weight') = (\weight - \weight')^\transpose \Lambda^{-1} (\weight - \weight')$, with $\Lambda = \mathrm{diag}(l^2_i), i=1,\dots,\featureDim$, as a length-scales matrix for automatic relevance determination (ARD). The same GP hyper-parameters adaptation scheme proposed in \cite{Wang2013} was applied to all BO methods. In the case of the GP noise model, since the simulations of the physics engine in the game are deterministic, we set the noise variance, $\sigma_n^2$ to 0.

As acquisition function for BO we applied the upper confidence bound (UCB) criterion:
\begin{equation}
h(\weights|\mathcal{D}) = \mu(\weights) + \beta \sigma(\weights),
\end{equation}
where $\mu(\weights)$ is the mean of the GP posterior at $\weights$, $\sigma(\weights)$ is the square root of the GP posterior variance, and $\beta$ is a parameter controlling the exploration-exploitation trade off. In most of our experiments, we were able to obtain good performance results with $\beta \in [0.5,2]$ using \methodShortName{}. For the experiments with REMBO, we used the Expected Improvement acquisition function, since the results on \cite{Wang2013} were achieved using this function.

The initial policy demonstration is given by a PI controller, whose only task is to drive the car at a constant speed of 15 m/s along the track, in all the experiments. An example of such data for one of the test tracks is shown in Figure \ref{fig:lengthscales}. That raw data is then fit through the specified number of policy kernels, yielding the initial set of weights $\weights_0$. All methods under comparison are informed with this initial solution to start the optimisation.

\begin{figure}
\centering
\subfloat[Forza]{\includegraphics[scale=0.15]{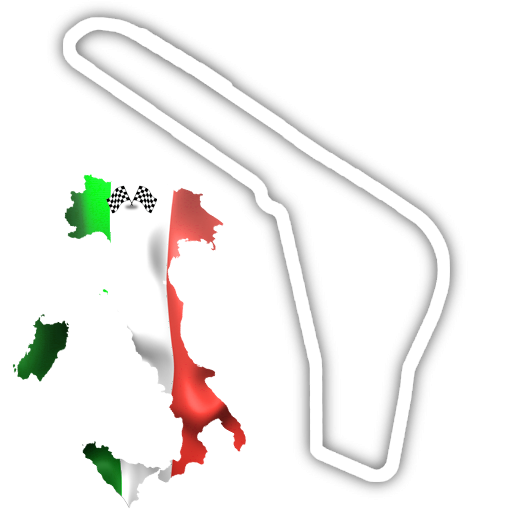}\label{fig:forza-outline}}
\subfloat[Allondaz]{\includegraphics[scale=0.15]{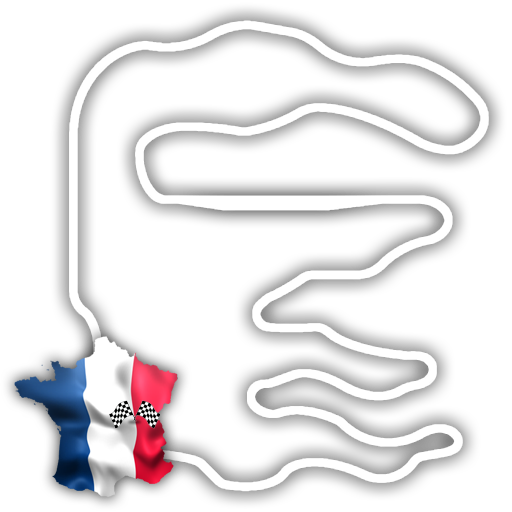}
\label{fig:allondaz-track}}
\caption{The race tracks for the experiments (Source: \textit{Speed Dreams})}
\label{fig:monza}
\end{figure}

\subsection{Results}
We tested the algorithms on two different tracks. Each one of them was given a budget of 300 policy evaluations. Each policy evaluation corresponds to one lap. In the case of BO and REMBO, the first 10 laps correspond to the initial samples (Section \ref{sec:algo}). Before each policy evaluation, to optimise the acquisition function, all versions of BO were allowed a maximum of 50,000 acquisition function evaluations, and had as starting point the best weights found up to that lap. To minimise randomness effects in the algorithms, each run of 300 laps was repeated 4 times, and the results were averaged.

The first track, called \textit{Forza} by the game, was a relatively simple circuit inspired by a real race track in Monza, Italy. This track is 5,784 metres long and 11 metres wide over flat asphalted terrain. The critical parts of this track are the sharp curves at the bottom of Figure \ref{fig:monza}a, which in terms of $\tp$ value, happen around 40\% of the track.

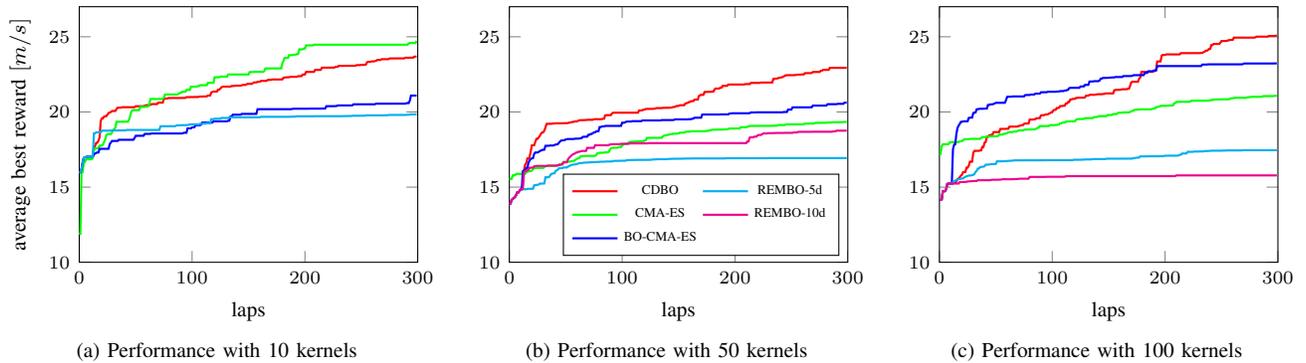
\begin{figure*}
\centering
\subfloat[Performance with 10 kernels]{
\tikzsetnextfilename{Forza-10}
\begin{tikzpicture}
\begin{axis}[xlabel={laps},ylabel={average best reward $[m/s]$},xmin=0, xmax=300, ymin=10, x=0.015cm, y=0.2cm, ymax=27, 
legend style={at={(0.95,0.05)},anchor=south east}, legend columns=1, transpose legend]
\addplot[red, solid, thick] table[header=false,x expr=\coordindex,y index=0] {data/forza2/bo-cd-10-average_best_reward.dat};
\addplot[green, solid, thick] table[header=false,x expr=\coordindex,y index=0] {data/forza2/cmaes-10-average_best_reward.dat};
\addplot[blue, solid, thick] table[header=false,x expr=\coordindex,y index=0] {data/forza2/bo-plain-10-average_best_reward.dat};
\addplot[cyan, solid, thick] table[header=false,x expr=\coordindex,y index=0] {data/forza/bo-rembo-10-average_best_reward.dat};
\end{axis}
\end{tikzpicture}
}
\subfloat[Performance with 50 kernels]{
\tikzsetnextfilename{Forza-50}
\begin{tikzpicture}
\begin{axis}[xlabel={laps},ylabel={},xmin=0, xmax=300, ymin=10, x=0.015cm, y=0.2cm, ymax=27, 
legend style={at={(0.975,0.025)},anchor=south east, font=\fontsize{5}{6}\selectfont}, legend columns=3, transpose legend]
\addplot[red, solid, thick] table[header=false,x expr=\coordindex,y index=0] {data/forza2/bo-cd-50-average_best_reward.dat};
\addlegendentry{\methodShortName{}}
\addplot[green, solid, thick] table[header=false,x expr=\coordindex,y index=0] {data/forza2/cmaes-50-average_best_reward.dat};
\addlegendentry{CMA-ES}
\addplot[blue, solid, thick] table[header=false,x expr=\coordindex,y index=0] {data/forza2/bo-plain-50-average_best_reward.dat};
\addlegendentry{BO-CMA-ES}
\addplot[cyan, solid, thick] table[header=false,x expr=\coordindex,y index=0] {data/forza/bo-rembo-50-average_best_reward.dat};
\addlegendentry{REMBO-5d}
\addplot[magenta, solid, thick] table[header=false,x expr=\coordindex,y index=0] {data/forza/bo-rembo-10d-50-average_best_reward.dat};
\addlegendentry{REMBO-10d}
\end{axis}
\end{tikzpicture}
}
\subfloat[Performance with 100 kernels]{
\tikzsetnextfilename{Forza-100}
\begin{tikzpicture}
\begin{axis}[xlabel={laps},ylabel={},xmin=0, xmax=300, ymin=10, x=0.015cm, y=0.2cm, ymax=27, 
legend style={at={(0.95,0.05)},anchor=south east, font=\fontsize{5}{6}\selectfont}, legend columns=3, transpose legend]
\addplot[red, solid, thick] table[header=false,x expr=\coordindex,y index=0] {data/forza2/bo-cd-100-average_best_reward.dat};
\addplot[green, solid, thick] table[header=false,x expr=\coordindex,y index=0] {data/forza2/cmaes-100-average_best_reward.dat};
\addplot[blue, solid, thick] table[header=false,x expr=\coordindex,y index=0] {data/forza2/bo-plain-100-average_best_reward.dat};
\addplot[cyan, solid, thick] table[header=false,x expr=\coordindex,y index=0] {data/forza/bo-rembo-100-average_best_reward.dat};
\addplot[magenta, solid, thick] table[header=false,x expr=\coordindex,y index=0] {data/forza/bo-rembo-10d-100-average_best_reward.dat};
\end{axis}
\end{tikzpicture}
}
\caption{Performance comparison w.r.t. dimensionality of the optimisation problem, evaluated on the \textit{Forza} track.}
\label{fig:forza-results}
\end{figure*}

Figure \ref{fig:forza-results} presents the overall performance of the analysed methods on \textit{Forza} for policies with different numbers of kernels. 
These experimental results allow us to assess how each method handles the increase in dimensionality, which adds flexibility to the control policy, but makes its optimisation harder. As we can see, although finding better solutions in the 10 kernels setting, CMA-ES's performance severely degrades with the increase in dimensionality. When combined with BO (BO-CMA-ES), CMA-ES helps it to improve performance on average, but what we observed across individual runs is that this behaviour is actually bimodal: sometimes very good, and other times very bad. REMBO wasn't able to achieve good results for any number of kernels in this track. For the 100 kernel test, one of the problems with REMBO is clearly visible: if the used random embedding is not able to capture a (or if there isn't any) relevant subspace, the performance is poor. Overall, it is possible to see that \methodShortName{}'s performance remains relatively stable with the increase in dimensionality and it is able to find better solutions for this policy search problem.

\begin{figure}
\subfloat[Control policy]{
	\tikzsetnextfilename{best-Forza}
	\begin{tikzpicture}
	\begin{axis}[
	xlabel={$\tp$},
	ylabel={$\action$},
	y=1.5cm,ymax=1,ymin=-0.1,xmin=-0.05,xmax=1,
	legend style={at={(0.7,0.95)},anchor=north east}, legend columns=2
	]
	\addplot[black,dashed] table[header=false,x index=0,y index=1, each nth point=50, filter discard warning=false, unbounded coords=discard, skip first n=10] {data/forza/best-50/initial_actions.dat};
	\addlegendentry{initial}
	\addplot[black,solid,thick] table[header=false,x index=0,y index=1, skip first n=10, each nth point=5, filter discard warning=false, unbounded coords=discard] {data/forza/best-50/played_actions.dat};
	\addlegendentry{$\policy^*$}
	\end{axis}
	\end{tikzpicture}
}

\subfloat[Resulting speeds]{
	\tikzsetnextfilename{speeds-Forza}
	\begin{tikzpicture}
	\begin{axis}[
	xlabel={$\tp$},
	ylabel={speed [m/s]},
	xmin=-0.05,xmax=1,
	y=0.04cm,
	legend style={at={(0.95,0.05)},anchor=south east},
	legend columns=2
	]
	\addplot[black,dashed,thick] table[header=false,x index=0,y index=2, each nth point=20, filter discard warning=false, unbounded coords=discard] {data/forza/best-50/initial_actions.dat};
	\addlegendentry{initial}
	\addplot[black,solid,thick] table[header=false,x index=0,y index=2, each nth point=5, filter discard warning=false, unbounded coords=discard] {data/forza/best-50/played_actions.dat};
	\addlegendentry{$\policy^*$}
	\end{axis}
	\end{tikzpicture}
}
\caption{Best policy obtained for track \textit{Forza}} 
\label{fig:forza-best}
\end{figure}
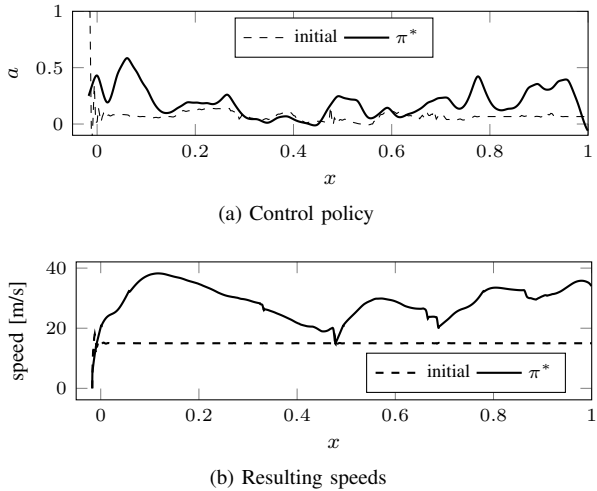

Figure \ref{fig:forza-best} presents the best policy with 50-kernels obtained by our method and the resulting speeds the car achieved along the \textit{Forza} track. From Figure \ref{fig:forza-best}b, it is possible to see that the algorithm adapts itself to speed up on the straight portions of the track and reduce speed close to curves, reducing the lap time.

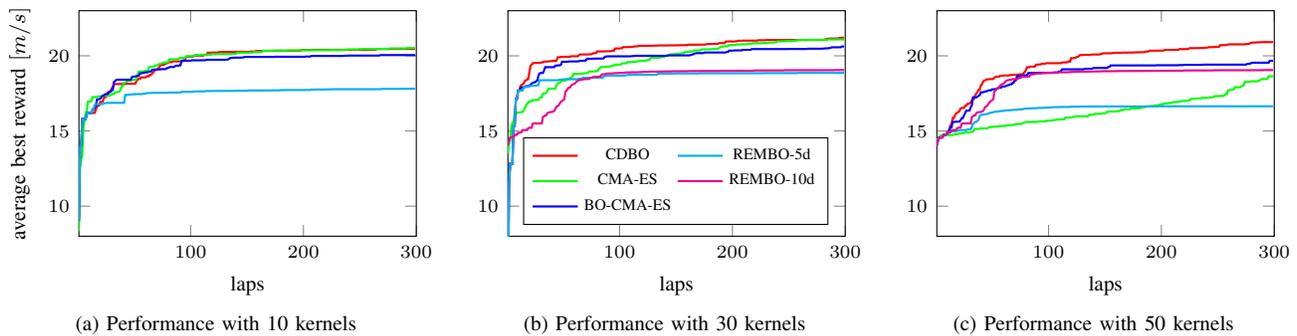
\begin{figure*}
\centering
\subfloat[Performance with 10 kernels]{
\tikzsetnextfilename{Allondaz-10}
\begin{tikzpicture}
\begin{axis}[xlabel={laps},ylabel={average best reward $[m/s]$},xmin=1, xmax=300, ymin=8, x=0.015cm, y=0.2cm, ymax=23, 
legend style={at={(0.95,0.05)},anchor=south east, font=\fontsize{6}{7}\selectfont}, legend columns=3, transpose legend]
\addplot[red, solid, thick] table[header=false,x expr=\coordindex,y index=0] {data/allondaz/bo-cd-10-average_best_reward.dat};
\addplot[green, solid, thick] table[header=false,x expr=\coordindex,y index=0] {data/allondaz/cmaes-10-average_best_reward.dat};
\addplot[blue, solid, thick] table[header=false,x expr=\coordindex,y index=0] {data/allondaz/bo-plain-10-average_best_reward.dat};
\addplot[cyan, solid, thick] table[header=false,x expr=\coordindex,y index=0] {data/allondaz/bo-rembo-10-average_best_reward.dat};
\end{axis}
\end{tikzpicture}
}
\subfloat[Performance with 30 kernels]{
\tikzsetnextfilename{Allondaz-30}
\begin{tikzpicture}
\begin{axis}[xlabel={laps},ylabel={},xmin=1, xmax=300, ymin=8, x=0.015cm, y=0.2cm, ymax=23, 
legend style={at={(0.95,0.05)},anchor=south east, font=\fontsize{6}{7}\selectfont}, legend columns=3, transpose legend]
\addplot[red, solid, thick] table[header=false,x expr=\coordindex,y index=0] {data/allondaz/bo-cd-30-average_best_reward.dat};
\addlegendentry{\methodShortName{}}
\addplot[green, solid, thick] table[header=false,x expr=\coordindex,y index=0] {data/allondaz/cmaes-30-average_best_reward.dat};
\addlegendentry{CMA-ES}
\addplot[blue, solid, thick] table[header=false,x expr=\coordindex,y index=0] {data/allondaz/bo-plain-30-average_best_reward.dat};
\addlegendentry{BO-CMA-ES}
\addplot[cyan, solid, thick] table[header=false,x expr=\coordindex,y index=0] {data/allondaz/bo-rembo-30-average_best_reward.dat};
\addlegendentry{REMBO-5d}
\addplot[magenta, solid, thick] table[header=false,x expr=\coordindex,y index=0] {data/allondaz/bo-rembo-10d-50-average_best_reward.dat};
\addlegendentry{REMBO-10d}
\end{axis}
\end{tikzpicture}
}
\subfloat[Performance with 50 kernels]{
\tikzsetnextfilename{Allondaz-50}
\begin{tikzpicture}
\begin{axis}[xlabel={laps},ylabel={},xmin=1, xmax=300, ymin=8, x=0.015cm, y=0.2cm, ymax=23, 
legend style={at={(0.95,0.05)},anchor=south east, font=\fontsize{6}{7}\selectfont}, legend columns=3, transpose legend]
\addplot[red, solid, thick] table[header=false,x expr=\coordindex,y index=0] {data/allondaz/bo-cd-50-average_best_reward.dat};
\addplot[green, solid, thick] table[header=false,x expr=\coordindex,y index=0] {data/allondaz/cmaes-50-average_best_reward.dat};
\addplot[blue, solid, thick] table[header=false,x expr=\coordindex,y index=0] {data/allondaz/bo-plain-50-average_best_reward.dat};
\addplot[cyan, solid, thick] table[header=false,x expr=\coordindex,y index=0] {data/allondaz/bo-rembo-50-average_best_reward.dat};
\addplot[magenta, solid, thick] table[header=false,x expr=\coordindex,y index=0] {data/allondaz/bo-rembo-10d-50-average_best_reward.dat};
\end{axis}
\end{tikzpicture}
}
\caption{Performance comparison w.r.t. dimensionality of the optimisation problem, evaluated on the track \textit{Allondaz}}
\label{fig:allondaz-results}
\end{figure*}

The second test track, \textit{Allondaz}, shown in Figure \ref{fig:allondaz-track}, is a road track with varying elevation along the path and filled with portions of complex geometry. It is 6,356 metres long and 12 metres wide, around the same dimensions as \textit{Forza}. The overall performance of the methods is presented in Figure \ref{fig:allondaz-results}. In this track no algorithm was able to achieve average speeds as high as in \textit{Forza}.  Despite increasing the number of kernels to optimise, the best performing policy, achieved with \methodShortName{}, does not significantly change when using beyond 50 kernels. That's why, for this track, we only present results for setups with up to 50 kernels. 

Similarly to the previous track, CMA-ES is not able to handle the high-dimensionality, and \methodShortName{} still maintains a consistent performance throughout the increase in dimensionality, demonstrating the capabilities of the method in high dimensions. Also, it's possible to see in all the tests that \methodShortName{} can achieve an even better result, if it is allowed to run for more iterations. REMBO achieved better results than the previous track, but it is still outperformed by BO-CMA-ES and CDBO. One interesting detail about REMBO's performance is that, after some iterations it does not improve any more, which means that it reached the optimum for the subspace used, and the global optimum is not in that subspace.

When compared to standard BO, another interesting feature of \methodShortName{} can be seen in Table \ref{table:runtime}, which shows the runtime for each experiment (300 laps) on \textit{Forza}. It is possible to see that with the increase in dimensionality, standard BO with CMA-ES (BO-CMA-ES) significantly increases in runtime when compared to the other methods. On the other hand, REMBO and our method maintain a low runtime through all the different problem dimensions. So, even if BO-CMA-ES achieves results close to \methodShortName{}, the runtime for the standard BO method is 10 times longer for the 100 dimensions case, which highlights the efficiency of the \methodShortName{} method for high-dimensional problems.

\begin{table}[t]
\centering
\caption{Average runtime (in seconds) on \textit{Forza}}
\begin{tabular}{|c|c|c|c|}
\hline
Method/Dimensions & 10 & 50 & 100\\
\hline
\methodShortName{} & 127 & 281 & 318\\
\hline
CMA-ES & 173 & 234 & 252\\
\hline
BO-CMA-ES & 553 & 680 & 3048\\
\hline
REMBO-5d & 154 & 242 & 257\\
\hline
REMBO-10d & - & 244 & 299\\
\hline
\end{tabular}
\label{table:runtime}
\end{table}

\section{Conclusion}
\label{sec:conclusion}

In this paper, we presented a method to optimise control policies to allow a robot to complete a given race track faster, which is an instantiation of a more general class of problems involving delayed rewards and costly policy evaluations. Our method applies Bayesian optimisation to guide the exploration of the parameter space towards the optimal policy. By making use of ideas from randomised coordinate descent methods, optimising a function one dimension at a time in a randomised sequence, and by starting the search from a valid initial solution, our method is able to be effective in the high-dimensional acquisition function optimisation sub-problem. Experiments with a car racing simulator demonstrated that this relatively simple approach is able to outperform other state-of-the-art black-box optimisation and BO methods in complex scenarios, some times in a fraction of the time. As future work, the model can be improved to deal with policies over higher-dimensional state and action spaces and to work together with other conventional motion planning algorithms.





%
%
%

\bibliographystyle{IEEEtran}

\bibliography{references}

\end{document}